\documentclass[11pt]{article}

\usepackage{acl} 
\usepackage{times}
\usepackage{latexsym}
\usepackage[T1]{fontenc}
\usepackage[utf8]{inputenc}
\usepackage{microtype}
\usepackage{graphicx}
\usepackage{booktabs}
\usepackage{amsmath}
\usepackage{algorithm}
\usepackage{algpseudocode}
\usepackage{tikz}
\usetikzlibrary{shapes, arrows, positioning, calc, fit}
\usepackage{float}
\usepackage{amssymb}
\usepackage{url}

\title{SG-UniBuc-NLP at SemEval-2026 Task 6: Multi-Head RoBERTa with Chunking for Long-Context Evasion Detection}

\author{Gabriel Stefan \and Sergiu Nisioi \\
  Human Language Technologies Research Center \\
  Faculty of Mathematics and Computer Science \\
  University of Bucharest \\
  \texttt{gabrielstefan04@gmail.com, sergiu.nisioi@unibuc.ro}}

\begin{document}
\maketitle

\begin{abstract}
We describe our system for SemEval-2026 Task 6 (CLARITY: Unmasking Political Question Evasions), which classifies English political interview responses by coarse-grained clarity (3-way) and fine-grained evasion strategy (9-way). Since responses frequently exceed the 512-token limit of standard Transformer encoders, we apply an overlapping sliding-window chunking strategy with element-wise Max-Pooling aggregation over chunk representations. A shared RoBERTa-large encoder supplies two task-specific heads trained jointly via a multi-task objective, with inference-time ensembling over 7-fold stratified cross-validation. Our system achieves a Macro-F1 of 0.80 on Subtask 1 and 0.51 on Subtask 2, ranking 11th in both subtasks. 
\end{abstract}

\section{Introduction}

Political interviews are a primary venue of democratic accountability, yet politicians routinely avoid giving direct answers. A meta-analysis of five televised interview studies by \citet{bull2003microanalysis} found that politicians provided clear responses to only 39--46\% of questions, compared to 70--89\% in non-political settings. This phenomenon, widely termed equivocation or evasion in the social-science literature 
\cite{harris1991,bull2003microanalysis,rasiah2010}, encompasses a range of rhetorical strategies, from subject shifts and partial answers to deliberate ambiguity, and has been the subject of detailed typological study \cite{clayman2001answers,bull1993how}. Despite this social-science foundation, the automatic detection of such strategies has received limited attention in NLP \cite{thomas2024isaidthatdataset}. 
SemEval-2026 Task~6 (\textit{CLARITY}: \textit{Unmasking Political Question Evasions}; \citealt{semeval2026task6}) addresses this gap by framing clarity and evasion detection as a supervised classification task over English political interview question--answer pairs, requiring systems to predict both a coarse-grained three-way \textit{Clarity} label and a fine-grained nine-way \textit{Evasion} label.

A central challenge is that political responses frequently exceed the 512-token input limit of standard Transformer encoders \cite{devlin2019bert,liu2019roberta,sanh2019distilbert}: na\"{\i}ve truncation risks discarding the precise span where evasion cues appear. We therefore segment each question--answer pair into overlapping 512-token chunks with a stride of 256 tokens, encode each chunk independently with a shared RoBERTa encoder, and aggregate the resulting chunk representations via element-wise Max-Pooling to obtain a single response-level vector \cite{pappagari2019hierarchical}. Two task-specific linear heads, a 3-way Clarity classifier and a 9-way Evasion classifier, are jointly trained on top of this shared encoder via a combined cross-entropy objective \cite{caruana1997multitask,ruder2017overview}, allowing the coarser clarity signal to regularize the more challenging evasion classification task. We train the entire system using 7-fold stratified cross-validation and combine the resulting fold models by averaging predicted probabilities at inference time \cite{galar2012review}.

On the official SemEval-2026 Task~6 evaluation, our ensemble achieves a Macro-F1 of 0.80 on Subtask~1 (Clarity), ranking 11th of 41 teams, and a Macro-F1 of 0.51 on Subtask~2 (Evasion), ranking 11th of 33 teams. Error analysis reveals two recurring failure modes driven by class imbalance and semantic overlap: \textit{Ambivalent} acts as a classification sink in Subtask~1, while performance on Subtask~2 severely degrades for minority categories with fine-grained pragmatic distinctions. Our implementation is publicly available.\footnote{\url{https://github.com/gabriel-stefan/political-evasion-detection}}

\section{Background}

\subsection{Task and Data}

SemEval-2026 Task~6 (\textit{CLARITY}; \citealt{semeval2026task6}) frames 
political evasion detection as a supervised classification problem over English question--answer (QA) pairs. Each instance consists of a question $Q$ and a response $A$; systems predict two labels simultaneously. \textbf{Subtask~1} requires 3-way \textit{Clarity} classification: \textit{Clear Reply} (question fully addressed), \textit{Clear Non-Reply} (speaker explicitly refuses), or \textit{Ambivalent} (response admits multiple interpretations). \textbf{Subtask~2} requires 9-way \textit{Evasion} classification into the taxonomy leaves of \citet{thomas2024isaidthatdataset}: \textit{Explicit}, \textit{Dodging}, \textit{Implicit}, \textit{General}, 
\textit{Deflection}, \textit{Partial/half-answer}, \textit{Clarification}, \textit{Claims ignorance}, or \textit{Declining to answer}. We participated in both subtasks; performance is evaluated with Macro-F1 \cite{semeval2026task6}.

\noindent\textbf{Example:}\\
$Q$: \textit{Based on your long experience, how does that change Finland's place in the world?}\\
$A$: \textit{Well, first of all, the context in which I said that was: The gentleman who occupies a seat\ldots}\\
Labels: \textit{Ambivalent} (Subtask~1), \textit{Dodging} (Subtask~2).

The CLARITY dataset \cite{thomas2024isaidthatdataset} comprises 3,756 English QA pairs from 287 official White House interview transcripts (2006--2023), split into 3,448 training and 308 development instances \cite{semeval2026task6}. Label distributions are substantially skewed: \textit{Ambivalent} constitutes 59\% of training instances, while \textit{Clear Non-Reply} accounts for only 10\%. At the evasion level, \textit{Explicit} (31\%) and \textit{Dodging} (21\%) dominate, while \textit{Partial/half-answer} (2\%), \textit{Clarification} (3\%), and \textit{Claims ignorance} (3\%), among others, are severely underrepresented. Inter-annotator agreement stands at Fleiss $\kappa = 0.64$ for Subtask~1 and $\kappa = 0.48$ for Subtask~2 \cite{thomas2024isaidthatdataset}, underscoring that fine-grained evasion 
categorization is challenging even for human judges. For development and test instances, multiple annotator labels are provided, and a prediction is scored as correct if it matches any of them \cite{semeval2026task6}.

\subsection{Related Work}

The CLARITY taxonomy builds on classical social-science typologies of 
political equivocation \cite{harris1991,bull2003microanalysis,rasiah2010}, 
consolidated into a computational label set by \citet{thomas2024isaidthatdataset}.

From a modeling perspective, the task poses a long-input classification 
challenge, as responses frequently exceed standard Transformer encoder limits. Dedicated long-context encoders such as Longformer \cite{beltagy2020longformer} and BigBird \cite{zaheer2020bigbird} address this via sparse attention, and more recent encoder-only models extend context further still \cite{warner-etal-2025-smarter}. An alternative line of work represents documents hierarchically by encoding fixed-length segments and aggregating their representations \cite{dietterich1997mil,ilse2018attentionmil, pappagari2019hierarchical}, which preserves full-document evidence while remaining compatible with standard pretrained encoders. 

\section{System Overview}
\label{sec:system}

\subsection{Hierarchical Input Processing}
\label{sec:chunking}
A key challenge in CLARITY is that responses can span thousands of tokens, exceeding the 512-token input limit of standard pretrained encoders such as RoBERTa \cite{liu2019roberta}. Since evasion cues may appear anywhere in the response, na\"{\i}ve truncation can remove critical evidence. We therefore adopt a sliding-window chunking strategy.

\begin{figure}[H]
    \centering
    \includegraphics[width=\columnwidth]{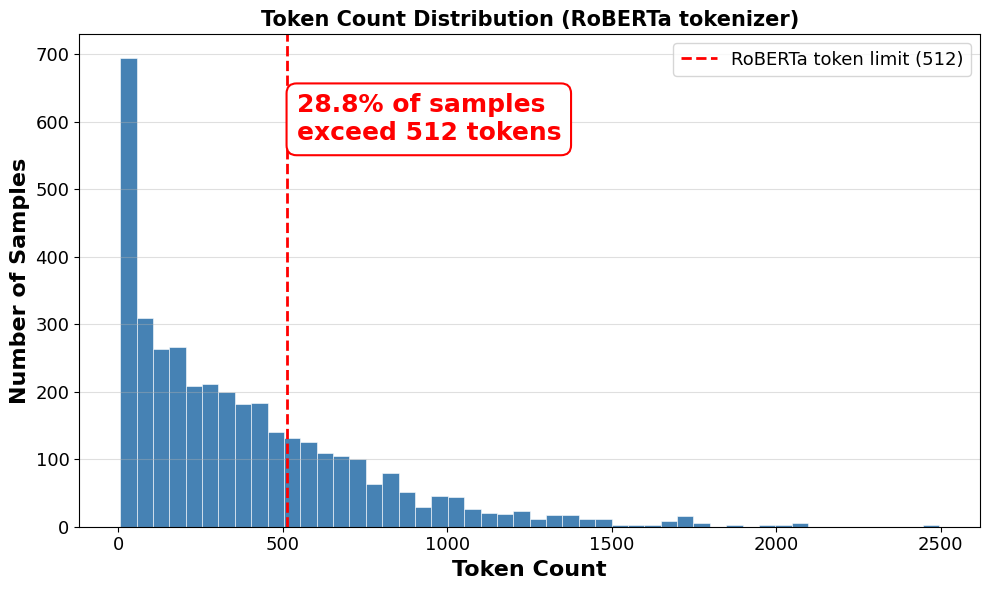}
    \caption{Distribution of token counts for the concatenated question--answer input sequence in the CLARITY dataset. The vertical red dashed line marks the 512-token limit of standard RoBERTa models.}
    \label{fig:token_dist}
\end{figure}

We adopt hierarchical chunking over long-context architectures (e.g., Longformer~\cite{beltagy2020longformer}, BigBird~\cite{zaheer2020bigbird}) primarily for memory efficiency, as chunking strictly bounds the memory footprint and prevents out-of-memory errors during fine-tuning. Additionally, this approach is straightforward to implement, as it integrates directly with any pretrained encoder without architectural modifications. Finally, our data distribution justifies this choice: since only 28.8\% of instances exceed 512 tokens (Figure~\ref{fig:token_dist}), the computational overhead of sparse-attention mechanisms is unnecessary for most inputs.

Given a question $Q$ and answer $A$, we construct a single input string using explicit prefixes:
\texttt{Question: \{Q\}\textbackslash nAnswer: \{A\}}.
Let $T$ denote the tokenized sequence of this full concatenation. Since $|T|$ can exceed the encoder limit, we segment $T$ into overlapping windows of length $L=512$ with stride $S=256$:
\begin{equation}
    C_k = T[kS : kS + L] \quad \text{for } k=0,\ldots,M-1,
\end{equation}
where $M = \left\lceil \frac{\max(|T|-L,0)}{S} \right\rceil + 1$ is the number of chunks for the instance. The final chunk always extends to $|T|$ and is padded to length $L$ if shorter. The window length $L=512$ is fixed by the encoder's maximum positional embedding capacity and is not a tunable hyperparameter. The stride $S=256$ was set as a principled 50\% overlap rather than through grid search: this overlap ensures that every token (except those in single-chunk documents) appears in at least two consecutive windows, reducing the risk that a chunk boundary splits a semantically coherent evasion span. 

Each chunk $C_k$ is encoded independently with a shared RoBERTa-large encoder. We extract the hidden state at position~0 of each chunk as a fixed-size chunk representation:

\begin{equation}
    h_k = H_k[0,:] \in \mathbb{R}^{d},
\end{equation}
where $H_k \in \mathbb{R}^{L \times d}$ is the final hidden state matrix for chunk $k$ and $d=1024$ for RoBERTa-large. Note that only the first chunk ($k{=}0$) begins with the actual \texttt{<s>} special token; for subsequent chunks, position~0 corresponds to the first token of that window. 

We then aggregate all chunk vectors for the same instance using element-wise Max-Pooling:
\begin{equation}
    v_j = \max_{k=0}^{M-1} h_{k,j} \quad \text{for } j=1,\ldots,d,
    \label{eq:maxpool}
\end{equation}
yielding a single response-level representation $v \in \mathbb{R}^d$.

\begin{algorithm}[h]
\caption{Long-Input Encoding Pipeline: overlapping chunking, per-chunk RoBERTa encoding, and element-wise Max-Pooling aggregation into a single response vector $v$.}
\label{alg:chunking}
\begin{algorithmic}[1]
\State \textbf{Input:} token sequence $T$ (length $|T|$), max length $L=512$, stride $S=256$
\State \textbf{Output:} response vector $v \in \mathbb{R}^d$
\State $chunks \gets [\ ]$
\State $start \gets 0$
\While{$start < |T|$}
    \State $end \gets \min(start + L,\ |T|)$
    \State $C \gets T[start : end]$
    \State pad $C$ to length $L$ and build attention mask
    \State $chunks.append(C)$
    \If{$end \ge |T|$}
        \State \textbf{break} \Comment{last chunk reaches end}
    \EndIf
    \State $start \gets start + S$
\EndWhile
\ForAll{$C_k \in chunks$} \Comment{encoded in a single batched forward pass}
    \State $H_k \gets \text{RoBERTa-large}(C_k)$
    \State $h_k \gets H_k[0,:]$ \Comment{position-0 embedding}
\EndFor
\State $v \gets \text{ElementWiseMax}(\{h_k\}_{k})$
\State \Return $v$
\end{algorithmic}
\end{algorithm}

\subsection{Multi-Task Learning Heads}
\begin{figure}[h]
\centering
\resizebox{\columnwidth}{!}{
\begin{tikzpicture}[
    node distance=0.8cm,
    block/.style={rectangle, draw, fill=blue!10, text width=2.5cm, text centered, rounded corners, minimum height=1cm},
    line/.style={draw, ->, >=latex, thick},
    chunk/.style={rectangle, draw, dashed, fill=gray!10, text width=1.5cm, minimum height=0.6cm, text centered}
]

\node (input) [text width=3.8cm, text centered] {Tokenized sequence \\ $T=\mathrm{tok}(Q \oplus A)$};

\node (chunk2) [chunk, below=0.5cm of input] {Chunk $1$};
\node (chunk1) [chunk, left=0.2cm of chunk2] {Chunk $0$};
\node (chunk3) [chunk, right=0.2cm of chunk2] {Chunk $M{-}1$};

\node (roberta) [rectangle, draw, fill=orange!20, minimum width=6.5cm, minimum height=0.8cm, below=0.6cm of chunk2]
{\textbf{Shared RoBERTa Encoder}};

\node (emb2) [below=0.6cm of roberta] {$h_1$};
\node (emb1) [left=1.5cm of emb2] {$h_0$};
\node (emb3) [right=1.5cm of emb2] {$h_{M-1}$};

\node (pool) [block, fill=green!10, below=0.6cm of emb2, text width=3.5cm]
{\textbf{Max-Pooling} \\ $v = \max_{k=0}^{M-1} h_k$};

\node (dropout) [block, fill=yellow!10, below=0.4cm of pool, text width=2.5cm] {Dropout};

\node (head_clarity) [block, fill=red!10, below left=0.8cm and -0.2cm of dropout]
{\textbf{Clarity} \\ (3 Classes)};
\node (head_evasion) [block, fill=red!10, below right=0.8cm and -0.2cm of dropout]
{\textbf{Evasion} \\ (9 Classes)};

\draw[line] (input) -- (chunk2);

\draw[dashed] (chunk1.south) -- (chunk1.south |- roberta.north);
\draw[dashed] (chunk2.south) -- (chunk2.south |- roberta.north);
\draw[dashed] (chunk3.south) -- (chunk3.south |- roberta.north);

\draw[line] (roberta.south -| emb1) -- (emb1);
\draw[line] (roberta.south -| emb2) -- (emb2);
\draw[line] (roberta.south -| emb3) -- (emb3);

\draw[line] (emb1) -- (pool);
\draw[line] (emb2) -- (pool);
\draw[line] (emb3) -- (pool);

\draw[line] (pool) -- (dropout);
\draw[line] (dropout) -| (head_clarity);
\draw[line] (dropout) -| (head_evasion);

\end{tikzpicture}
}
\caption{System architecture. The tokenized concatenated input ($Q\oplus A$) is split into overlapping chunks, each chunk is encoded by a shared RoBERTa encoder, chunk representations are aggregated via element-wise Max-Pooling, and two task-specific heads predict Clarity (3-way) and Evasion (9-way).}
\label{fig:architecture}
\end{figure}
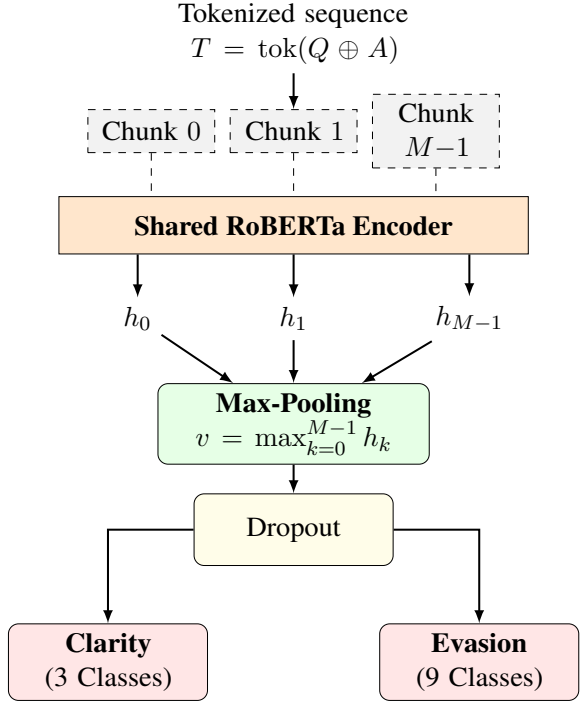

As shown in Figure~\ref{fig:architecture}, the pooled vector $v$ (Eq.~\ref{eq:maxpool}) is shared by both subtasks. We apply dropout \cite{srivastava2014dropout} ($p=0.1$) to $v$ and 
feed it into two task-specific linear heads: a 3-way classifier for Clarity and a 9-way classifier for Evasion.

\begin{align}
    \hat{y}_c &= \mathrm{softmax}(W_c \cdot \mathrm{Dropout}(v) + b_c), \\
    \hat{y}_e &= \mathrm{softmax}(W_e \cdot \mathrm{Dropout}(v) + b_e),
\end{align}
where $W_c \in \mathbb{R}^{3 \times d}$ and $W_e \in \mathbb{R}^{9 \times d}$.

We train both heads jointly using standard cross-entropy losses with equal weighting:
\begin{equation}
\mathcal{L} = \mathcal{L}_{c} + \mathcal{L}_{e}.
\end{equation}

We investigate alternative loss functions in Section~\ref{sec:results}.

\subsection{Training and Inference}
\label{sec:training}

We train using 7-fold stratified cross-validation, stratifying folds by the Subtask~1 (Clarity) labels to preserve class proportions. In each fold, we select the checkpoint that maximizes the combined validation score (defined below in Section~\ref{sec:metrics}), treating both subtasks equally. Full training hyperparameters are listed in Table~\ref{tab:hyperparams}.

At inference time, we ensemble all 7 fold models by averaging their 
predicted class probabilities and taking $\arg\max$:
\begin{equation}
    \hat{y} = \arg\max_{c} \;\frac{1}{7}\sum_{i=1}^{7} p^{(i)}_c,
\end{equation}
where $p^{(i)}_c$ is the predicted probability for class $c$ from fold 
model $i$.

\section{Experimental Setup}
\label{sec:experiments}

\subsection{Data Splits}
We follow the official SemEval-2026 Task~6 setup \cite{semeval2026task6}. 
The organizers provide a labeled \texttt{train} split, a labeled \texttt{dev} split (released as \texttt{test} during the competition but used here as the official development set) and a blind \texttt{test} set. We train models on \texttt{train} and  estimate performance via 7-fold stratified cross-validation (Section~\ref{sec:training}). 

\subsection{Preprocessing}
Each instance is formatted as a single sequence using the template \texttt{Question: \{Q\}\textbackslash nAnswer: \{A\}}. No truncation is applied prior to chunking; the full concatenated sequence is tokenized and passed directly to the sliding-window pipeline described in Section~\ref{sec:chunking}. Sequences are dynamically padded to the batch maximum length by the data collator; within the model, individual chunks shorter than $L{=}512$ tokens are zero-padded to $L$.

We fine-tune RoBERTa-large \cite{liu2019roberta} using AdamW \cite{loshchilov2019adamw} with a linear learning-rate schedule and warmup; exact training hyperparameters and configuration details are provided in Appendix~\ref{app:hyperparams}.

\subsection{Evaluation Metrics}
\label{sec:metrics}
Following the task guidelines \cite{semeval2026task6}, the primary metric is Macro-F1 for both subtasks. Macro-F1 computes the unweighted average of per-class F1 scores. For early stopping and checkpoint selection during cross-validation, we compute the combined score:
\begin{equation}
    \text{F1}_{\text{comb}} = \frac{1}{2}\left(\text{F1}_{\text{clarity}} 
    + \text{F1}_{\text{evasion}}\right),
\end{equation}
which treats both subtasks equally and avoids optimizing for one at the expense of the other.

\subsection{Implementation}
We implement our system in Python 3.12 using PyTorch and Hugging Face Transformers (full dependencies provided in Appendix~\ref{app:dependencies}). Random seeds are fixed to 42 for reproducibility. All experiments were conducted on a single NVIDIA RTX 3090 (24\,GB VRAM). Training the full 7-fold ensemble takes approximately 5 hours.

\section{Results and Analysis}
\label{sec:results}

\subsection{Main Results}
On the official SemEval-2026 Task~6 blind test set (submitted under the CodaBench username \texttt{gabriel-stefan}), our final ensemble achieves a Macro-F1 of 0.80 on Subtask~1 (Clarity), ranking 11th out of 41 teams. On Subtask~2 (Evasion), the system achieves a Macro-F1 of 0.51, ranking 11th out of 33 participating teams.

\subsection{Ablation Studies}
To isolate the impact of individual design choices, each ablation varies one component while keeping the rest fixed to our full configuration: hierarchical Max-Pooling, multi-task training, and 7-fold stratified cross-validation. We report mean and standard deviation across folds.

\paragraph{Pooling Strategy}
Table~\ref{tab:pooling} shows Max-Pooling outperforms Mean-Pooling and the First-Chunk baseline. Mean-Pooling averages representations, potentially diluting localized signals. Conversely, element-wise Max-Pooling extracts the maximum feature activations across all chunks, creating a composite representation that likely preserves the strongest evasion cues regardless of position, explaining the improved fine-grained recall.

\begin{table}[H]
\centering
\small
\begin{tabular}{lcc}
\toprule
\textbf{Pooling Method} & \textbf{Clarity F1} & \textbf{Evasion F1} \\
\midrule
First Chunk Only & $0.67 \pm 0.01$ & $0.42 \pm 0.01$ \\
Mean Pooling & $0.68 \pm 0.02$ & $0.43 \pm 0.02$ \\
\textbf{Max-Pooling (Ours)} & $\mathbf{0.70 \pm 0.02}$ & $\mathbf{0.45 \pm 0.02}$ \\
\bottomrule
\end{tabular}
\caption{Max-Pooling outperforms Mean-Pooling and the First-Chunk baseline across both subtasks (7-fold CV; mean $\pm$ std).}
\label{tab:pooling}
\end{table}

\paragraph{Multi-Task Learning}
Table~\ref{tab:multitask} compares joint Multi-Task training against single-task models, demonstrating the benefit of auxiliary supervision. While Clarity performance remains constant, the Multi-Task objective improves Evasion Macro-F1 from $0.42$ to $0.45$. This suggests high-level clarity labels provide an effective regularization signal for the more complex evasion task.

\begin{table}[H]
\centering
\resizebox{\columnwidth}{!}{%
\small
\begin{tabular}{lcc}
\toprule
\textbf{Training Objective} & \textbf{Clarity F1} & \textbf{Evasion F1} \\
\midrule
Single-task (Clarity only) & $\mathbf{0.70 \pm 0.02}$ & -- \\
Single-task (Evasion only) & -- & $0.42 \pm 0.01$ \\
\textbf{Multi-task (Ours)} & $0.70 \pm 0.02$ & $\mathbf{0.45 \pm 0.02}$ \\
\bottomrule
\end{tabular}%
}
\caption{Effect of multi-task learning (7-fold CV; mean $\pm$ std). Joint training improves the more difficult Evasion task via auxiliary regularization.}
\label{tab:multitask}
\end{table}

\paragraph{Ensemble Size}
To quantify the trade-off between robustness and computational cost, we trained three variants using stratified $k$-fold cross-validation with $k\in\{3,5,7\}$ (stratified by Subtask~1 labels). For each $k$, we report the mean $\pm$ standard deviation of validation Macro-F1 across the $k$ held-out folds. At test time, each variant ensembles the $k$ fold models by averaging their predicted probabilities. As shown in Table~\ref{tab:ensemble_tradeoff}, increasing $k$ yields consistent gains, especially for the more difficult Evasion subtask, at the cost of proportionally higher training and inference time.

\begin{table}[H]
\centering
\resizebox{\columnwidth}{!}{%
\small
\begin{tabular}{lccc}
\toprule
\textbf{Ensemble size} & \textbf{Clarity F1} & \textbf{Evasion F1} 
    & \textbf{Rel.\ cost} \\
\midrule
3-fold  & $0.66 \pm 0.01$ & $0.42 \pm 0.02$ & $3.0\times$ \\
5-fold  & $0.68 \pm 0.02$ & $0.43 \pm 0.03$ & $5.0\times$ \\
\textbf{7-fold (Ours)} & $\mathbf{0.70 \pm 0.02}$ & 
    $\mathbf{0.45 \pm 0.02}$ & $7.0\times$ \\
\bottomrule
\end{tabular}%
}
\caption{Increasing ensemble size consistently improves both subtasks at proportionally higher cost, with each larger $k$ also providing more training data per fold model (7-fold CV; mean $\pm$ std).}
\label{tab:ensemble_tradeoff}
\end{table}

\paragraph{Class Imbalance Mitigation}
To evaluate alternative strategies for handling class imbalance (Section~\ref{sec:training}), we trained two additional variants: inverse-frequency class-weighted cross-entropy and focal loss ($\gamma=2$; \citealt{lin2017focal}). Table~\ref{tab:imbalance_ablation} reports the 7-fold cross-validation results.

\begin{table}[h]
\centering
\small
\begin{tabular}{lcc}
\toprule
\textbf{Loss function} & \textbf{Clarity F1} & \textbf{Evasion F1} \\
\midrule
Cross-entropy (ours)    & $0.70 \pm 0.02$ & $0.45 \pm 0.02$ \\
Class-weighted CE       & $0.66 \pm 0.02$ & $0.41 \pm 0.02$ \\
Focal loss ($\gamma=2$) & $0.69 \pm 0.02$ & $0.44 \pm 0.02$ \\
\bottomrule
\end{tabular}
\caption{Effect of loss function on class imbalance (7-fold CV,
mean $\pm$ std).}
\label{tab:imbalance_ablation}
\end{table}

Neither alternative improves Macro-F1, but error patterns reveal a meaningful redistribution of performance rather than uniform degradation. Class weighting increases minority-class recall: \textit{Clear Non-Reply} rises from 0.62 to 0.67, and \textit{Partial/half-answer} increases from 0.00 to 0.10 on out-of-fold predictions (the only setting predicting it at all). However, these gains are eclipsed by majority-class precision drops (\textit{Ambivalent} F1 falls from 0.78 to 0.75; \textit{Explicit} F1 from 0.65 to 0.60), resulting in a net decrease in the equally-weighted Macro-F1. Focal loss falls between these extremes, avoiding sharp majority-class degradation but yielding only marginal minority-class recovery.

We attribute this to two compounding factors. First, data sparsity is the binding constraint: with only 79 \textit{Partial/half-answer} instances, reweighting amplifies gradients but cannot compensate for lacking lexical and pragmatic diversity. The 10\% recall confirms the model can detect this strategy, but the available signal remains too sparse and noisy. Second, dominant confusions reflect semantic overlap, not sampling bias. The largest error clusters involve pragmatically adjacent classes (\textit{Implicit}, \textit{Deflection}, \textit{General}, \textit{Dodging}) sharing surface features that confound even human annotators (Fleiss $\kappa = 0.48$). Reweighting amplifies gradients without resolving this representational ambiguity, potentially destabilizing the shared encoder with conflicting supervision for examples with similar surface forms. These findings suggest improving minority classes requires targeted data augmentation rather than loss engineering (Section~\ref{sec:conclusion}).

\subsection{Error Analysis}
\label{sec:error_analysis}

We analyze out-of-fold predictions on the 3{,}448 training instances (Table~\ref{tab:per_class_stats}; confusion matrices in Appendix~\ref{app:confusion}).

\begin{table}[ht]
\centering
\small
\begin{tabular}{lrcc}
\toprule
\textbf{Class} & \textbf{n} & \textbf{Acc} & \textbf{Conf} \\
\midrule
\multicolumn{4}{l}{\textit{Subtask 1 (Clarity)}} \\
Ambivalent       & 2040 & 0.784 & 0.936 \\
Clear Non-Reply  &  356 & 0.643 & 0.927 \\
Clear Reply      & 1052 & 0.623 & 0.935 \\
\midrule
\multicolumn{4}{l}{\textit{Subtask 2 (Evasion)}} \\
Clarification        &   92 & 0.707 & 0.892 \\
Explicit             & 1052 & 0.643 & 0.830 \\
Declining to answer  &  145 & 0.607 & 0.838 \\
Claims ignorance     &  119 & 0.521 & 0.795 \\
Dodging              &  706 & 0.474 & 0.757 \\
General              &  386 & 0.306 & 0.710 \\
Implicit             &  488 & 0.281 & 0.706 \\
Deflection           &  381 & 0.228 & 0.676 \\
Partial/half-answer  &   79 & 0.000 & 0.705 \\
\bottomrule
\end{tabular}
\caption{Per-class accuracy and overall mean prediction confidence (Conf) from out-of-fold predictions on the training set. Note that Conf is computed across all instances (both correct and incorrect) for a given class.}
\label{tab:per_class_stats}
\end{table}

\paragraph{Subtask 1 (Clarity)}
While Table~\ref{tab:per_class_stats} reports overall confidence, we observe that the model's mean prediction confidence remains remarkably high even when isolating exclusively the misclassified instances (\textit{Ambivalent}: 0.898, \textit{Clear Reply}: 0.911, \textit{Clear Non-Reply}: 0.888). Due to severe imbalance, \textit{Ambivalent} (59.2\% of train) acts as a majority-class sink, absorbing 35.6\% of \textit{Clear Reply} and 30.6\% of \textit{Clear Non-Reply} instances, whereas only 16.9\% of \textit{Ambivalent} cases are misclassified as \textit{Clear Reply}. This asymmetry suggests our unweighted loss biases the model to treat \textit{Ambivalent} as a high-probability prior for borderline cases.

\paragraph{Subtask 2 (Evasion)}
Lexical surface cues clearly differentiate performance across categories. Formulaic classes (\textit{Clarification}: 70.7\%, \textit{Explicit}: 64.3\%, \textit{Declining to answer}: 60.7\%) outperform implicature-reliant categories (\textit{General}: 30.6\%, \textit{Implicit}: 28.1\%, \textit{Deflection}: 22.8\%) and exhibit higher mean confidence (0.835 vs 0.698, frequency-weighted). Notably, \textit{Partial/half-answer} ($n=79$) yields zero recall, with predictions scattered across seven classes. Furthermore, \textit{Implicit} is frequently misclassified as \textit{Explicit} (28.1\%), while \textit{Deflection} splits between \textit{Explicit} (20.7\%) and \textit{Dodging} (18.9\%), reflecting semantic overlap among non-refusal redirections.

\paragraph{Model Errors and Annotator Disagreement}
On the dev set, model errors correlate strongly with human disagreement. Evasion majority-vote agreement averages 44.0\% overall but diverges sharply by consensus: 58.4\% on unanimous items ($n=125$) versus 32.0\% on 2-1 splits ($n=150$). While any-annotator agreement reaches 59.1\%, this lenient metric inflates amid maximally ambiguous 1-1-1 cases. Subtask 1 exhibits a parallel trend, with clarity accuracy dropping from 80.0\% (unanimous evasion cases) to 68.7\% (2-1 splits). These results indicate that a substantial fraction of model errors occurs on inherently ambiguous instances where annotator consensus is weak.

\paragraph{Qualitative Analysis}
The qualitative examples in Appendix~\ref{app:qual_examples} reveal a shared failure mode: over-reliance on surface forms, where assertive language masks evasive pragmatics. These failures highlight that simple input concatenation fails to force the model to verify if the response actually resolves the question's core demand.

\section{Conclusion}
\label{sec:conclusion}
We presented a hierarchical multi-task system for SemEval-2026 Task~6 (CLARITY) designed for long-context political interview responses. Our approach combines overlapping sliding-window chunking with Max-Pooling aggregation to preserve full-response evidence, and jointly trains two classification heads. Our ensemble achieved a Macro-F1 of 0.80 on Subtask 1 and 0.51 on Subtask 2, ranking 11th in both subtasks. 

Max-Pooling consistently outperformed Mean-Pooling and the first-chunk baseline, confirming that evasion cues are often localized. Error analysis revealed that inherent ambiguity remains challenging: performance degrades on subtle minority categories often confused with explicit or implicit answers, whereas strategies with strong lexical cues maintain high recall.

Future work will explore targeted augmentation for minority classes. Additionally, question--answer concatenation may fail to capture the relational structure underlying evasion. We therefore plan to investigate Natural Language Inference (NLI) frameworks and dual-encoder late-interaction models such as ColBERT~\cite{khattab2020colbert}, which allow questions to query token-level answer representations.

\section*{Acknowledgments}
We thank the SemEval-2026 Task 6 organizers for the CLARITY dataset and evaluation platform, and the anonymous reviewers for their valuable feedback that improved this work.

This research is supported by the project ``Romanian Hub for Artificial Intelligence - HRIA'', Smart Growth, Digitization and Financial Instruments Program, 2021-2027, MySMIS no. 351416.

\bibliography{main} 

\appendix

\section{Training Hyperparameters}
\label{app:hyperparams}

\begin{table}[H]
\centering
\resizebox{\columnwidth}{!}{%
\small
\begin{tabular}{ll}
\toprule
\textbf{Hyperparameter} & \textbf{Value} \\
\midrule
Optimizer & AdamW (weight decay 0.01) \\
Learning rate & $1\mathrm{e}{-5}$ (10\% warmup) \\
Batch size & 8 \\
Max epochs & 15 (early stopping patience = 3) \\
Classifier dropout & 0.1 \\
Gradient clipping & max norm 1.0 \\
Precision & BF16 mixed precision \\
Gradient checkpointing & enabled \\
Random seed & 42 (base; per-fold offset) \\
\bottomrule
\end{tabular}%
}
\caption{Training configuration and hyperparameters.}
\label{tab:hyperparams}
\end{table}

\section{Implementation Details and Dependencies}
\label{app:dependencies}

Our system was developed and evaluated using the following libraries and specific versions:

\vspace{0.5em}
{\raggedright
\texttt{torch} ($\ge$2.2.2): \url{https://pytorch.org} \\
\texttt{transformers} ($\ge$4.40.0): \url{https://huggingface.co/docs/transformers} \\
\texttt{datasets} ($\ge$2.19.0): \url{https://huggingface.co/docs/datasets} \\
\texttt{accelerate} ($\ge$0.30.0): \url{https://huggingface.co/docs/accelerate} \\
\texttt{scikit-learn} ($\ge$1.4.2): \url{https://scikit-learn.org} \\
\texttt{numpy} ($\ge$1.26.4): \url{https://numpy.org} \\
\texttt{pandas} ($\ge$2.2.2): \url{https://pandas.pydata.org} \\
\texttt{protobuf} (==3.20.3): \url{https://protobuf.dev} \\
\texttt{sentencepiece} ($\ge$0.2.0): \url{https://github.com/google/sentencepiece}
\par}

\section{Confusion Matrices}
\label{app:confusion}

In this section, we provide the row-normalized confusion matrices for both subtasks to visualize the recall performance across classes.

\begin{figure}[H]
    \centering
    % Clarity Matrix
    \includegraphics[width=1\columnwidth]{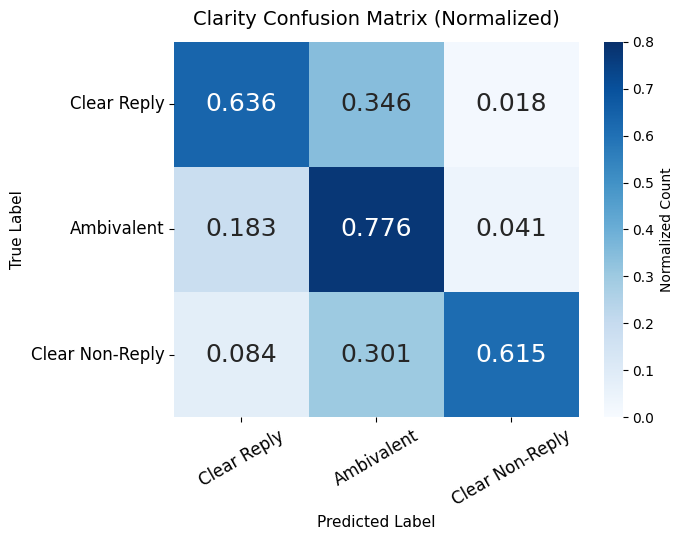}
    \caption{Normalized Confusion Matrix for Subtask 1 (Clarity)}
    \label{fig:cm_clarity}
\end{figure}

\begin{figure}[H]
    \centering
    \includegraphics[width=1\columnwidth]{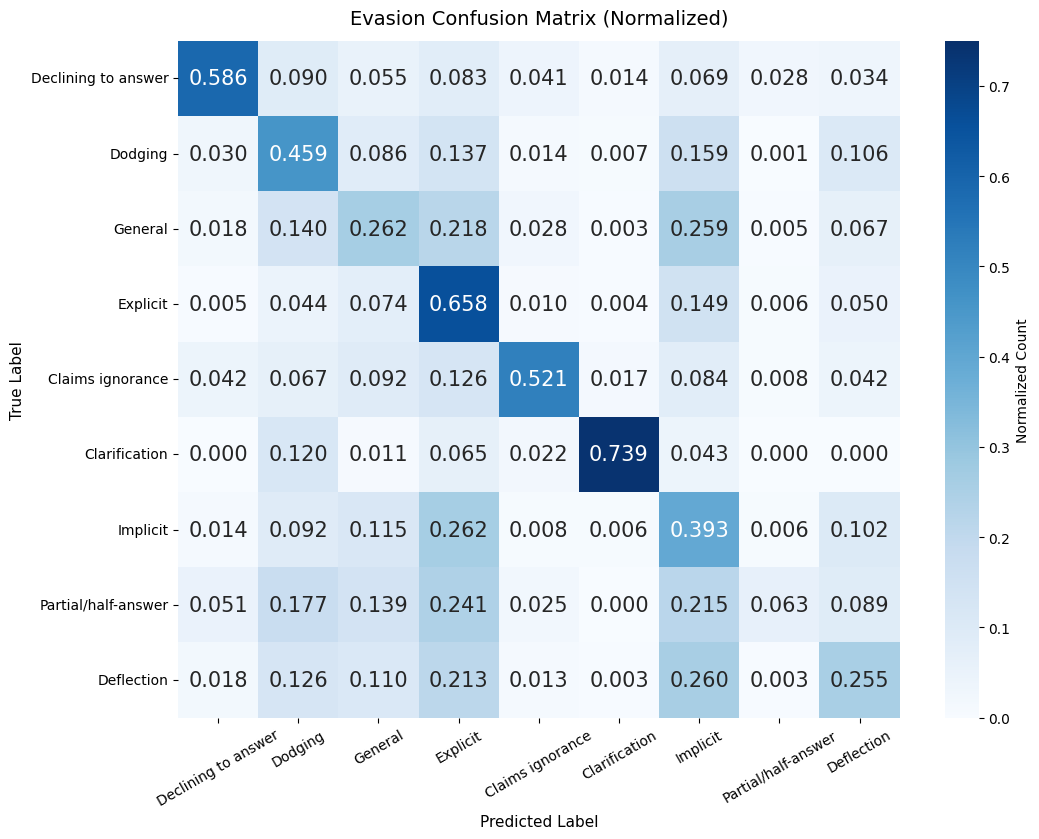}
    \caption{Normalized Confusion Matrix for Subtask 2 (Evasion)}
    \label{fig:cm_evasion}
\end{figure}

\section{Qualitative Error Examples}
\label{app:qual_examples}

Table~\ref{tab:qual_examples} exemplifies the error patterns in Section~\ref{sec:error_analysis}, specifically the model's reliance on surface lexical cues over pragmatic inference. While explicit meta-refusals (Example 1) enable correct, high-confidence (0.990) predictions of \textit{Clear Non-Reply} and \textit{Declining to answer}, more subtle strategies often lead to misclassification. In Example 2, the model incorrectly predicts \textit{Explicit} for a \textit{Partial/half-answer}, prioritizing a direct-sounding response about timing while overlooking the omitted location. Similarly, in Example 3, assertive first-person surface features override the pragmatic intent of ironic humor, causing a \textit{Deflection} to be misread as an \textit{Explicit Clear Reply}. These instances reinforce that simple concatenation fails to force the model to verify if responses resolve the question's core demand.

\clearpage
\onecolumn
\begin{table}[H]
\centering
\small
\begin{tabular}{p{5.8cm} p{5.0cm} cc cc}
\toprule
\textbf{Question (abridged)} & \textbf{Answer (abridged)} &
\multicolumn{2}{c}{\textbf{Gold}} &
\multicolumn{2}{c}{\textbf{Predicted}} \\
\cmidrule(lr){3-4}\cmidrule(lr){5-6}
& & \textbf{Clar.} & \textbf{Eva.} & \textbf{Clar.} & \textbf{Eva.} \\
\midrule
\textit{Do you think it was the right action for Israel to take?} &
\textit{I'm not going to comment on the subject that you're trying to get me to comment on.} &
CNR & DTA & CNR & DTA \\
\addlinespace
\textit{Your acceptance speech --- are you physically going to be in Charlotte, or will you give the speech here?} &
\textit{We'll be doing a speech on Thursday --- the main speech. Charlotte --- they will be doing nominations on Monday. I speak on Thursday.} &
AMB & PHA & AMB & EXP \\
\addlinespace
\textit{Do you think that if there is a breach, nobody is going to blame you?} &
\textit{Of course, no one would blame me. I know you won't. You'll be saying, Biden did a wonderful job.} &
AMB & DEF & CR & EXP \\
\bottomrule
\end{tabular}
\caption{Representative prediction outcomes from out-of-fold analysis. Abbreviations---Clar.: Clarity; Eva.: Evasion; CNR: Clear Non-Reply; CR: Clear Reply; AMB: Ambivalent; DTA: Declining to answer; EXP: Explicit; PHA: Partial/half-answer; DEF: Deflection.}
\label{tab:qual_examples}
\end{table}

\end{document}